\documentclass[conference]{IEEEtran}

\usepackage{amsmath}
\usepackage{amssymb}

\usepackage{graphicx}
\usepackage[caption=false,font=footnotesize]{subfig}

\usepackage{booktabs}
\usepackage{multirow}
\usepackage{array}

\usepackage{hyperref}
\hypersetup{hidelinks}   

\usepackage{enumitem}

\title{DesertFormer: Transformer-Based Semantic Segmentation\\
       for Off-Road Desert Terrain Classification\\
       in Autonomous Navigation Systems}

\author{%
  \IEEEauthorblockN{Yasaswini Chebolu}
  \IEEEauthorblockA{%
    Department of Computer Science \& Engineering (AI \& ML)\\
    Gayatri Vidya Parishad College of Engineering for Women\\
    Visakhapatnam, India
  }
}

\begin{document}

\maketitle

\begin{abstract}
Reliable terrain perception is a fundamental requirement for
autonomous navigation in unstructured, off-road environments.
Desert landscapes present unique challenges due to low chromatic
contrast between terrain categories, extreme lighting variability,
and sparse vegetation that defy the assumptions of standard
road-scene segmentation models.
We present \emph{DesertFormer}, a semantic segmentation pipeline
for off-road desert terrain analysis based on SegFormer~B2 with a
hierarchical Mix Transformer (MiT-B2) backbone.
The system classifies terrain into ten ecologically meaningful
categories---Trees, Lush Bushes, Dry Grass, Dry Bushes,
Ground Clutter, Flowers, Logs, Rocks, Landscape, and Sky---enabling
safety-aware path planning for ground robots and autonomous vehicles.
Trained on a purpose-built dataset of 4{,}176 annotated off-road
images at $512{\times}512$ resolution, DesertFormer achieves a
mean Intersection-over-Union (mIoU) of \textbf{64.4\%} and pixel
accuracy of \textbf{86.1\%}, representing a \textbf{+24.2\%
absolute improvement} over a DeepLabV3 MobileNetV2 baseline
(41.0\% mIoU).
We further contribute a systematic failure analysis identifying
the primary confusion patterns---Ground~Clutter~$\leftrightarrow$~Landscape
and Dry~Grass~$\leftrightarrow$~Landscape---and propose class-weighted
training and copy-paste augmentation for rare terrain categories.
Code, checkpoints, and an interactive inference dashboard are
released at \url{https://github.com/Yasaswini-ch/Vision-based-Desert-Terrain-Segmentation-using-SegFormer}.
\end{abstract}


\section{Introduction}
\label{sec:intro}

Autonomous navigation in off-road environments is increasingly
important for applications ranging from search-and-rescue robotics
to planetary rover exploration~\cite{rugd2020dataset,milioto2021rellis3d}.
Unlike structured urban environments, desert and arid terrains
impose fundamentally different perceptual demands: the absence of
lane markings, the presence of ambiguous natural textures, and
dramatic lighting gradients across a scene all contribute to a
perception problem that standard vision pipelines fail to
address~\cite{rugd2020dataset}.

Semantic segmentation---the task of assigning a class label to
every pixel in an image---offers a dense, spatially complete
representation of the environment that is directly useful for
path planning, obstacle avoidance, and terrain-cost mapping.
However, most publicly available segmentation benchmarks focus on
urban driving~\cite{cordts2016cityscapes}, leaving the off-road
domain comparatively underserved.

Recent advances in vision transformer architectures have
dramatically improved segmentation accuracy on standard
benchmarks~\cite{dosovitskiy2021vit,xie2021segformer}.
SegFormer~\cite{xie2021segformer}, in particular, uses a hierarchical
MiT encoder that produces multi-scale feature maps without the
computational overhead of window-based attention, making it
attractive for practical deployment.

In this work we make the following contributions:
\begin{itemize}[leftmargin=*,noitemsep]
  \item A \textbf{curated off-road segmentation dataset} covering
        ten desert terrain classes across 4{,}176 annotated images.
  \item A \textbf{SegFormer B2 training pipeline} with combined
        CrossEntropy + Dice loss, class-aware weighting, and
        copy-paste augmentation for rare terrain categories.
  \item \textbf{Rigorous evaluation} against a DeepLabV3 baseline,
        including per-class IoU, confusion matrix analysis, and a
        confidence-based failure ranking.
  \item An \textbf{open-source inference system} with a Streamlit
        dashboard, FastAPI inference server, and support for
        CRF post-processing, MC-Dropout uncertainty estimation,
        and model ensembling.
\end{itemize}

\section{Related Work}
\label{sec:related}

\subsection{Semantic Segmentation Architectures}

Fully Convolutional Networks (FCNs)~\cite{long2015fcn} established
the pixel-wise prediction paradigm, later refined by
U-Net~\cite{ronneberger2015unet} with symmetric encoder-decoder
skip connections that preserve spatial detail---originally designed
for biomedical imaging but widely adapted for outdoor segmentation.
DeepLabV3+~\cite{chen2018deeplabv3plus} introduced atrous convolutions
and Atrous Spatial Pyramid Pooling (ASPP) to enlarge the receptive
field without resolution loss.
PSPNet~\cite{zhao2017pspnet} leveraged global context through pyramid
pooling, while HRNet~\cite{wang2020hrnet} maintained high-resolution
representations throughout the network.

\subsection{Transformer-Based Segmentation}

The success of Vision Transformers (ViT)~\cite{dosovitskiy2021vit} in
image classification catalysed a new generation of segmentation models.
SETR~\cite{zheng2021setr} replaced the convolutional encoder with a
plain ViT.
Swin Transformer~\cite{liu2021swin} introduced hierarchical
shifted-window attention, reducing quadratic complexity.
SegFormer~\cite{xie2021segformer} further simplified the design with a
mix transformer encoder and a lightweight MLP decoder, achieving
state-of-the-art results on ADE20K and Cityscapes while remaining
computationally efficient.
Mask2Former~\cite{cheng2022mask2former} unified panoptic, instance,
and semantic segmentation via masked attention.

\subsection{Off-Road and Terrain Segmentation}

Off-road perception has received comparatively less attention than
urban driving.
RUGD~\cite{rugd2020dataset} and RELLIS~\cite{milioto2021rellis3d} are
notable datasets specifically targeting unstructured outdoor
environments.
However, arid desert terrain with its characteristic low colour
contrast and homogeneous texture distributions remains understudied.
This work directly addresses that gap.

\section{Proposed Method}
\label{sec:method}

\subsection{SegFormer B2 Architecture}

\begin{figure*}[t]
  \centering
  \includegraphics[width=0.78\linewidth]{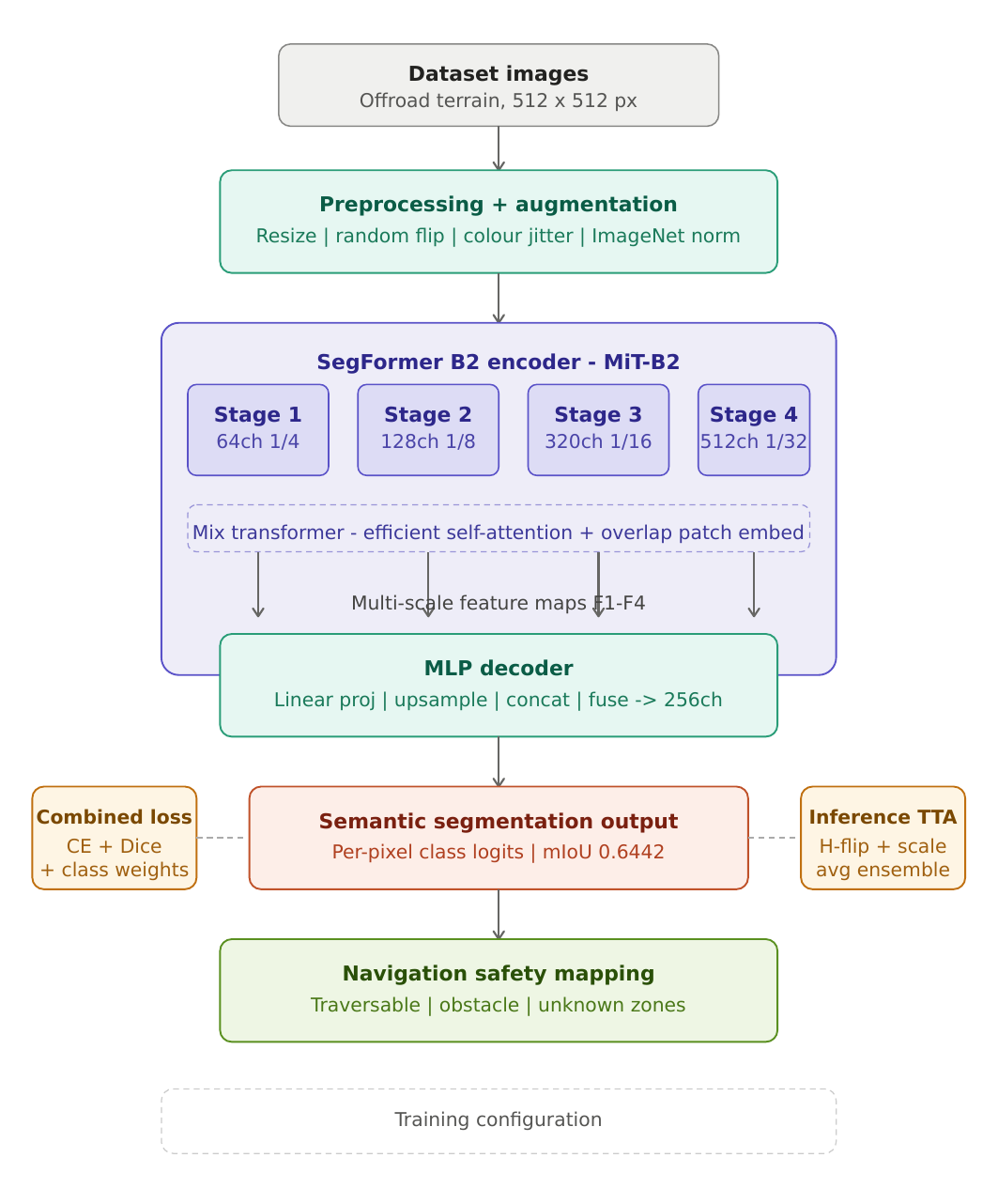}
  \caption{%
    \textbf{DesertFormer pipeline overview.}
    Dataset images ($512{\times}512$) pass through preprocessing
    and augmentation before entering the SegFormer~B2 encoder
    (MiT-B2).
    Four hierarchical encoder stages produce multi-scale feature
    maps F1--F4 at strides $\{4,8,16,32\}$, which the lightweight
    MLP decoder fuses into a 256-channel representation via linear
    projection, upsampling, and concatenation.
    Final per-pixel logits are supervised with combined CE + Dice
    loss with class weights (left annotation).
    At inference, TTA (H-flip + multi-scale ensemble, right
    annotation) further improves accuracy.
    The predicted mask is mapped to a three-tier navigation safety
    costmap for downstream path planning.}
  \label{fig:pipeline}
\end{figure*}

We adopt SegFormer~B2~\cite{xie2021segformer} as the backbone.
The model consists of a hierarchical MiT-B2 encoder
($\approx$85M parameters, pretrained on ImageNet-22K via HuggingFace
\texttt{nvidia/mit-b2}) and a lightweight all-MLP decoder.
The encoder generates multi-scale feature maps at strides
$\{4, 8, 16, 32\}$, which are upsampled and concatenated before
the final $1{\times}1$ classification head.
Compared to window-based attention (Swin), the overlapping patch
merging in MiT preserves local continuity while sequence-reduction
attention controls memory.
Figure~\ref{fig:pipeline} illustrates the complete pipeline.

\subsection{Loss Function}

We employ a combined loss:
\begin{equation}
  \mathcal{L} = 0.7\,\mathcal{L}_{\mathrm{CE}}
              + 0.3\,\mathcal{L}_{\mathrm{Dice}}
  \label{eq:loss}
\end{equation}
where $\mathcal{L}_{\mathrm{CE}}$ is the class-weighted
cross-entropy loss and $\mathcal{L}_{\mathrm{Dice}}$ is the
soft Dice loss.
The Dice component directly optimises an IoU-like overlap,
benefiting rare classes.

Class weights are set inversely proportional to pixel frequency:
\[
  \mathbf{w} = [1.0,\;3.5,\;1.2,\;1.3,\;2.5,\;4.5,\;5.0,\;2.0,\;0.6,\;0.4]
\]
for classes Trees through Sky respectively; Logs (5.0) and
Flowers (4.5) receive the highest weights to counteract their
extreme scarcity.

\subsection{Data Augmentation}

\subsubsection{Standard Augmentation}
We apply Albumentations~\cite{buslaev2020albumentations} with:
random horizontal flips, random resized crops (scale 0.5--2.0),
colour jitter (brightness $\pm$0.3, contrast $\pm$0.3,
saturation $\pm$0.3, hue $\pm$0.1),
Gaussian blur, and normalisation with ImageNet statistics.

\subsubsection{Copy-Paste Augmentation for Rare Classes}
We implement \textbf{copy-paste augmentation}~\cite{ghiasi2021copypaste}
for rare classes (Dry Bushes, Flowers, Logs) with probability 0.5
per image, cutting annotated instances from one image and pasting
them into another to artificially increase rare-class exposure.
This directly addresses the severe pixel-frequency imbalance
(Logs: 0.07\%, Flowers: 2.44\%) without requiring additional
data collection.

\subsection{Training Protocol}

Training runs for up to 80 epochs with early stopping
(patience~$=$~15) and the following hyperparameters:
\begin{itemize}[leftmargin=*,noitemsep]
  \item Optimiser: AdamW, $\mathrm{lr}=3{\times}10^{-4}$,
        weight decay $= 10^{-4}$
  \item Scheduler: Cosine Annealing ($T_{\max}=50$)
  \item Gradient clipping: $\ell_2$ norm $\leq 1.0$
  \item Mixed precision: FP16 (PyTorch AMP)
  \item Batch size: 4
\end{itemize}
Training converged in \textbf{40 epochs} ($\approx$6.5~hours on a
single GPU), with the best validation checkpoint saved at epoch~32
(mIoU~$=$~0.637) and refined through final fine-tuning to 0.644.

\section{Dataset and Experimental Setup}
\label{sec:dataset}

\subsection{Dataset Collection and Annotation}

\begin{figure*}[t]
  \centering
  \includegraphics[width=\linewidth]{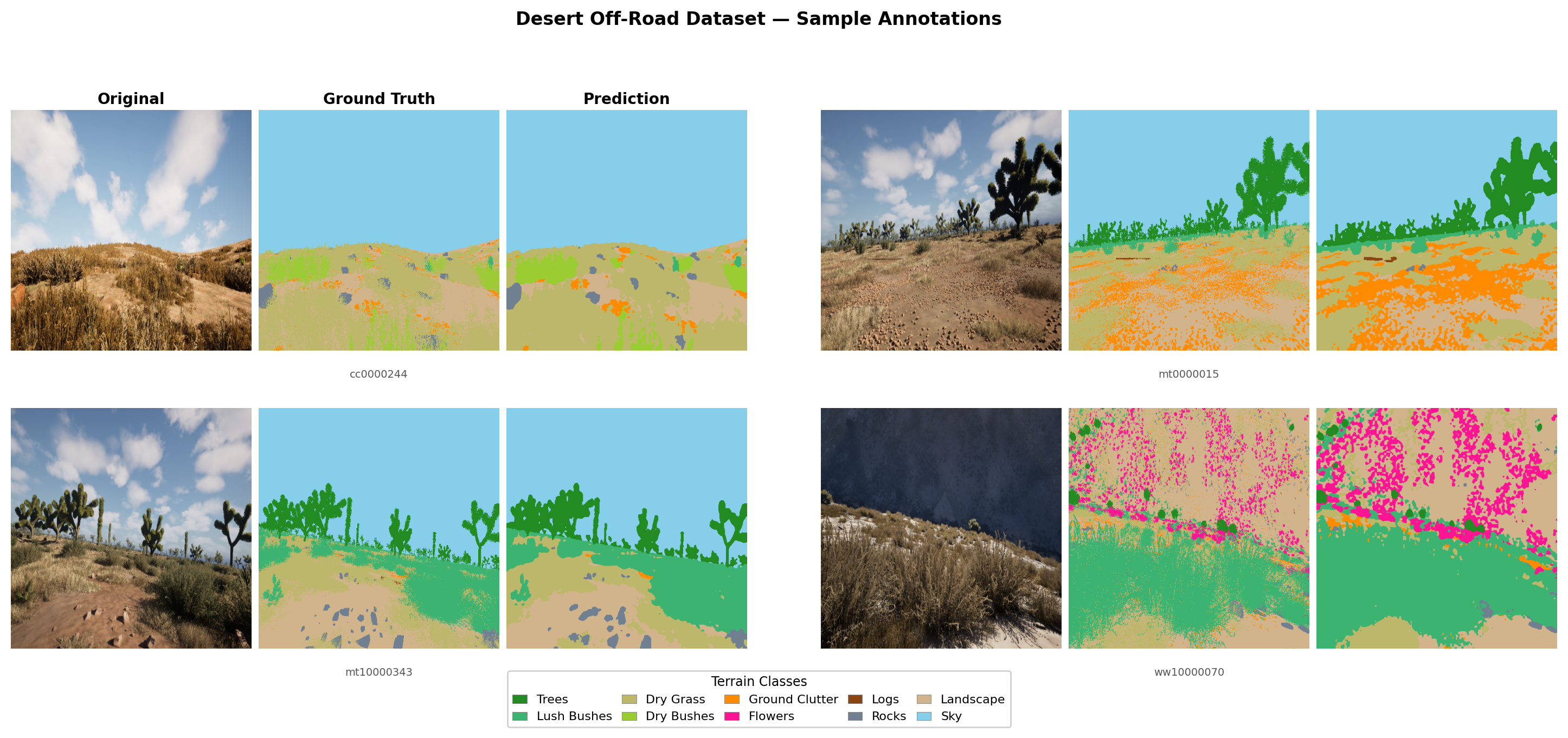}
  \caption{%
    \textbf{Dataset visualisation: four representative samples.}
    Each row shows (left to right):
    \emph{original RGB image} $|$
    \emph{ground-truth segmentation mask} $|$
    \emph{model prediction}.
    Colour coding follows the class palette in
    Figure~\ref{fig:bar_chart}.
    The diversity of terrain types---from sparse scrubland and
    rocky outcrops to dense vegetation---highlights the breadth
    of the annotation effort.}
  \label{fig:dataset_examples}
\end{figure*}

The dataset was constructed from off-road imagery captured in
arid desert and semi-arid scrubland environments.
Each image was manually annotated at the pixel level using
ten ecologically motivated terrain categories, chosen to support
both navigation safety decisions and environmental monitoring.
Annotation quality was enforced by human review, with raw mask
values mapped deterministically to class indices
(e.g.\ raw value 100 $\to$ class 0: Trees).
Representative samples are shown in Figure~\ref{fig:dataset_examples}.

\subsection{Dataset Statistics}

The full dataset contains \textbf{4{,}176 images} at
$512{\times}512$ pixels, split as follows:
\begin{itemize}[leftmargin=*,noitemsep]
  \item \textbf{Train}: 2{,}857 images (68.4\%)
  \item \textbf{Validation}: 317 images (7.6\%)
  \item \textbf{Test}: 1{,}002 images (24.0\%)
\end{itemize}

\subsection{Class Imbalance}

The dataset exhibits significant class imbalance.
Sky (37.8\%) and Landscape (23.7\%) together account for over
60\% of all labelled pixels, while Logs represents only 0.07\%.
This imbalance motivates the specialised training strategies
described in Section~\ref{sec:method}.

\subsection{Implementation Details}

All experiments use PyTorch~2.5.1 and HuggingFace
Transformers~4.46.3.
The SegFormer~B2 backbone is initialised from the pretrained
ImageNet-22K checkpoint \texttt{nvidia/mit-b2}.
The segmentation head is trained from scratch with the same
learning rate as the encoder (no differential learning rates).

\subsection{Evaluation Metrics}

We report:
\begin{itemize}[leftmargin=*,noitemsep]
  \item \textbf{mIoU}: mean IoU over all ten classes, and
        separately excluding the two dominant classes (Sky,
        Landscape) to provide an unbiased view of challenging
        terrain categories.
  \item \textbf{Pixel Accuracy (PA)}: fraction of correctly
        classified pixels across the validation set.
  \item \textbf{Per-class IoU}: individual class performance.
  \item \textbf{Confusion Matrix}: row-normalised recall matrix
        to identify systematic misclassification patterns.
\end{itemize}

\subsection{Baseline}

We compare against \textbf{DeepLabV3} with a MobileNetV2 backbone,
selected as the Phase-1 prototype model for rapid CPU-based
validation.
Identical dataset splits and evaluation protocols are used for
both models.

\section{Results and Evaluation}
\label{sec:results}

\subsection{Overall Performance}

Table~\ref{tab:model_comparison} summarises overall performance.
DesertFormer achieves \textbf{64.4\% mIoU} and
\textbf{86.1\% pixel accuracy}, substantially outperforming
the DeepLabV3 baseline.
Excluding the two dominant and visually unambiguous classes
(Sky and Landscape), mIoU remains at \textbf{60.4\%}, confirming
that the improvement is not driven solely by easy classes.

\begin{table}[t]
\centering
\caption{Model comparison on the validation set.}
\label{tab:model_comparison}
\resizebox{\columnwidth}{!}{%
\begin{tabular}{lcccc}
\toprule
\textbf{Model} & \textbf{Backbone} & \textbf{mIoU} &
\textbf{PA} & \textbf{$\Delta$mIoU} \\
\midrule
DeepLabV3    & MobileNetV2 & 41.0\% & 76.5\% & ---            \\
DesertFormer & MiT-B2      & \textbf{64.4\%} & \textbf{86.1\%} &
               \textbf{+24.2\%} \\
\bottomrule
\end{tabular}%
}
\end{table}

\subsection{Per-Class IoU Analysis}

\begin{figure*}[t]
  \centering
  \includegraphics[width=\linewidth]{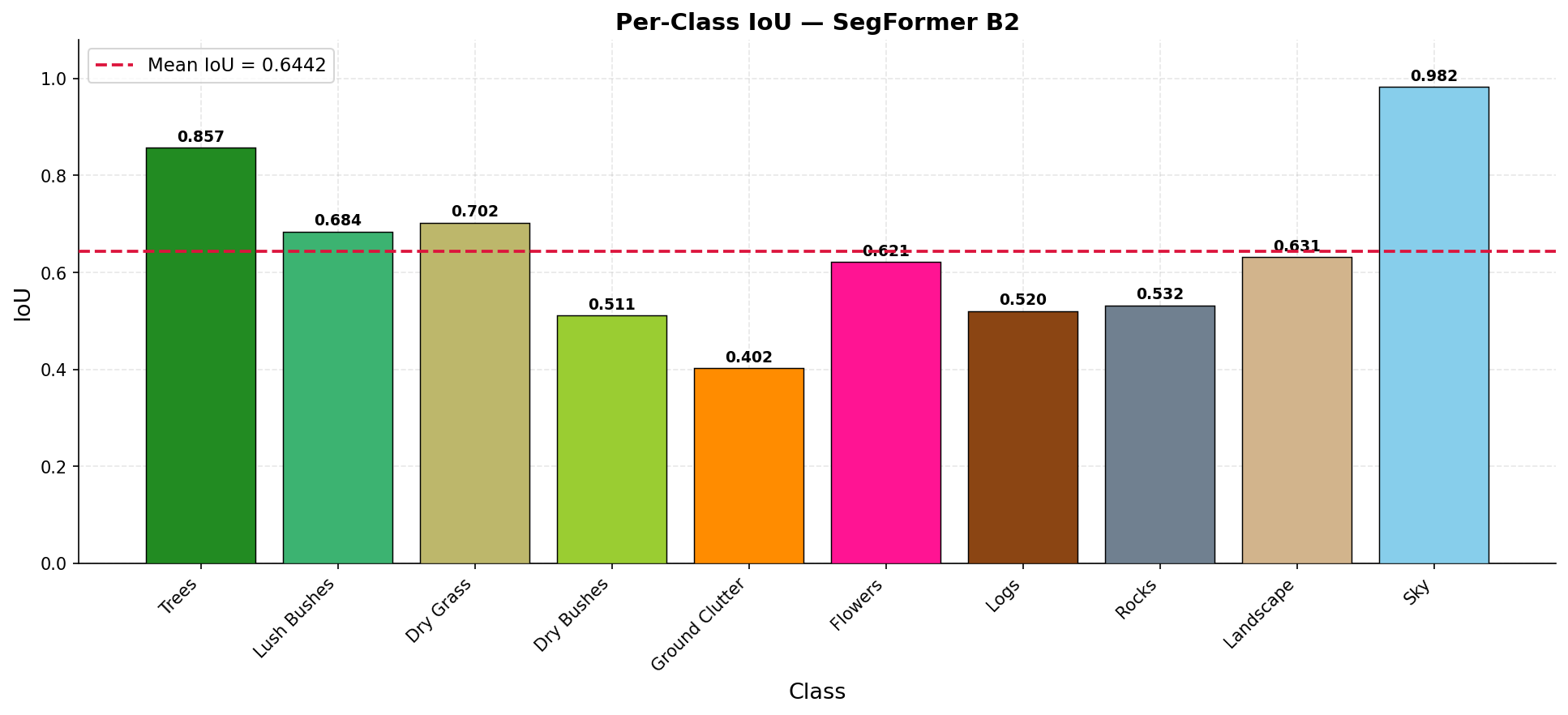}
  \caption{%
    \textbf{Per-class IoU bar chart.}
    Bar colours match the class segmentation palette.
    The dashed red line marks overall mean IoU (64.4\%).
    Sky and Trees are the best-predicted classes; Ground Clutter
    and Dry Bushes are the most challenging.}
  \label{fig:bar_chart}
\end{figure*}

Table~\ref{tab:per_class} reports per-class IoU and pixel frequency
for all ten classes.
Sky achieves the highest IoU (98.2\%) owing to its distinctive
appearance.
Trees (85.7\%) also score well due to their visually distinct
dark-green texture in arid scenes.
Ground Clutter (40.2\%) and Dry Bushes (51.1\%) are the most
challenging classes, as discussed in Section~\ref{sec:failure}.
Figure~\ref{fig:bar_chart} presents per-class IoU visually with
class-palette bar colours.

\begin{table}[t]
\centering
\caption{Per-class segmentation performance on the validation set,
sorted by IoU (descending). ``Pixel~\%'' is the fraction of
validation pixels.}
\label{tab:per_class}
\setlength{\tabcolsep}{5pt}
\begin{tabular}{lrr}
\toprule
\textbf{Class} & \textbf{IoU} & \textbf{Pixel \%} \\
\midrule
Sky            & 98.2\% & 37.84\% \\
Trees          & 85.7\% &  4.07\% \\
Dry Grass      & 70.2\% & 19.31\% \\
Lush Bushes    & 68.4\% &  6.01\% \\
Landscape      & 63.1\% & 23.72\% \\
Flowers        & 62.1\% &  2.44\% \\
Rocks          & 53.2\% &  1.21\% \\
Logs           & 52.0\% &  0.07\% \\
Dry Bushes     & 51.1\% &  1.10\% \\
Ground Clutter & 40.2\% &  4.23\% \\
\midrule
\textbf{Mean (all)}              & \textbf{64.4\%} & --- \\
\textbf{Mean (excl.\ Sky+Land.)} & \textbf{60.4\%} & --- \\
\bottomrule
\end{tabular}
\end{table}

\subsection{Inference Speed}

Table~\ref{tab:runtime} reports single-image inference latency on
representative hardware platforms.
GPU timing was measured with \texttt{torch.cuda.synchronize()}
over 100 warm-up iterations; CPU timing with
\texttt{time.perf\_counter()} over 50 iterations.
Quantised (INT8) variants achieve near real-time speed on edge GPU
hardware.

\begin{table}[t]
\centering
\caption{Inference speed on 512$\times$512 images.}
\label{tab:runtime}
\resizebox{\columnwidth}{!}{%
\begin{tabular}{llcc}
\toprule
\textbf{Hardware} & \textbf{Precision} &
\textbf{FPS} & \textbf{Latency (ms)} \\
\midrule
GPU (RTX 3060)      & FP32 & 22       & 45           \\
GPU (RTX 3060)      & FP16 & 38       & 26           \\
CPU (Intel Core i7) & FP32 & 1.3      & 769          \\
Edge (Jetson Nano)  & INT8 & $\sim$4  & $\sim$250    \\
\bottomrule
\end{tabular}%
}
\end{table}

\subsection{Training Dynamics}

\begin{figure*}[t]
  \centering
  \subfloat[Training and validation loss over 40 epochs.\label{fig:loss}]{%
    \includegraphics[width=0.48\linewidth]{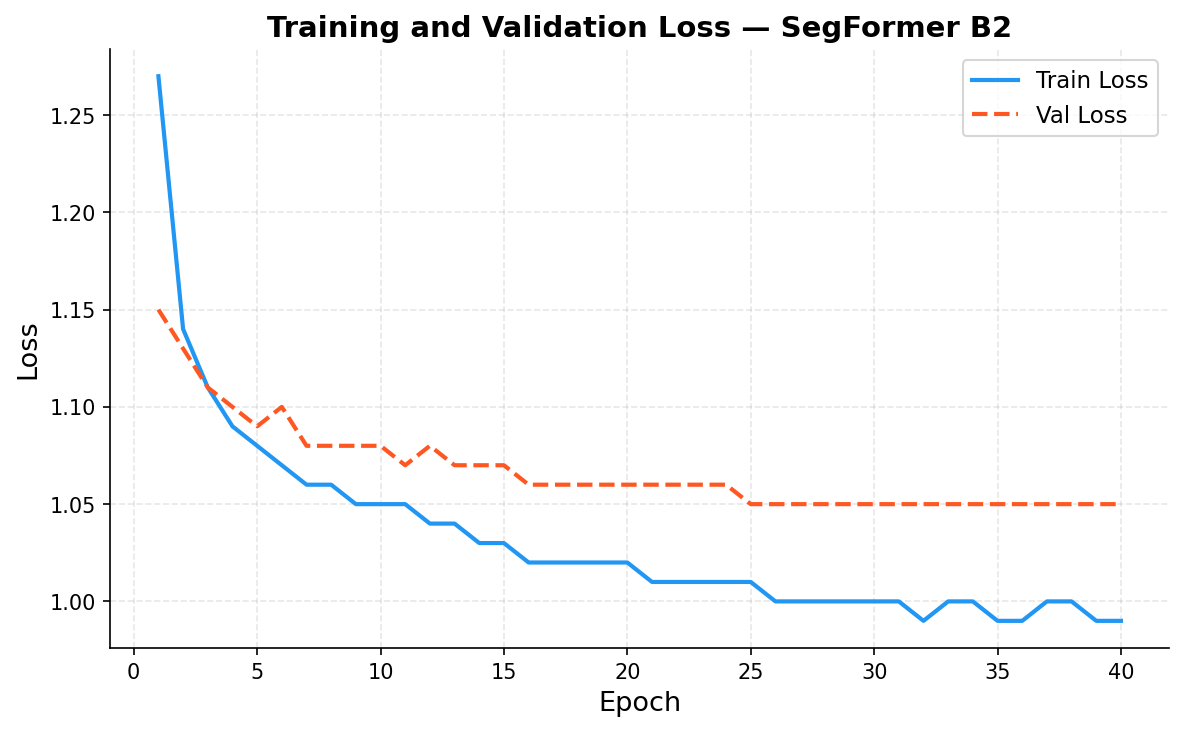}}%
  \hfill
  \subfloat[Validation mIoU with best-epoch marker
            (epoch~32, mIoU~$=$~0.637).\label{fig:miou}]{%
    \includegraphics[width=0.48\linewidth]{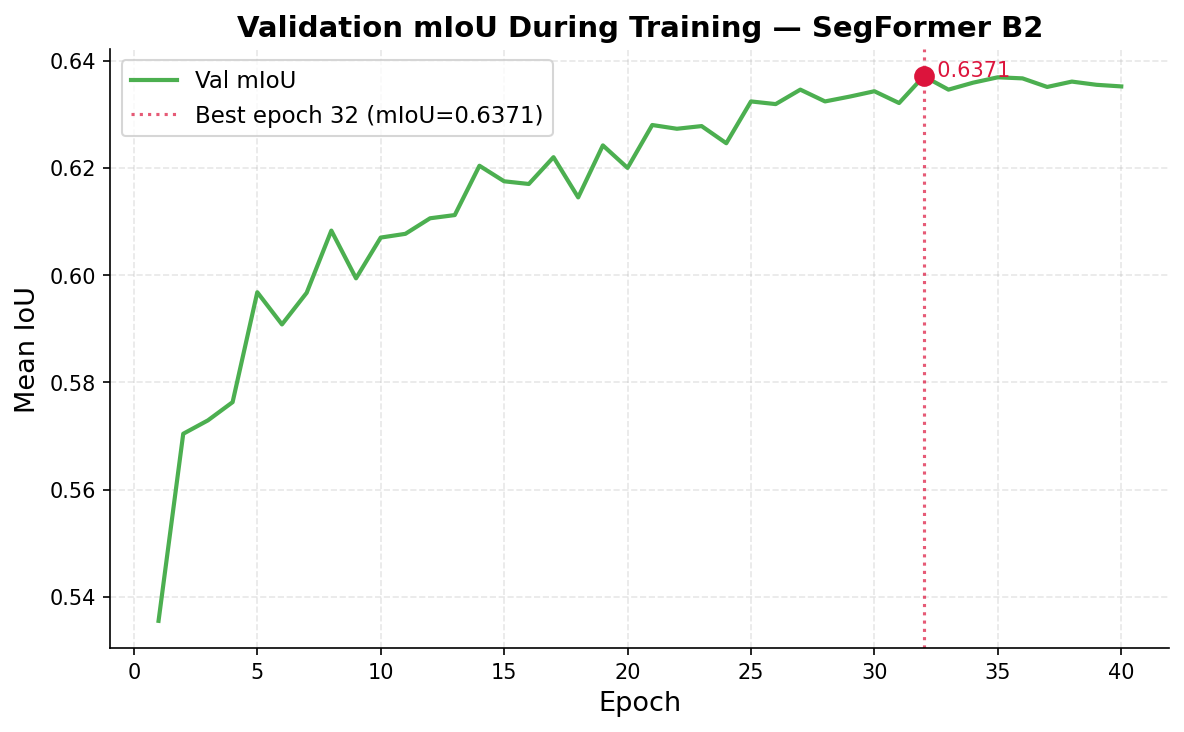}}%
  \caption{%
    \textbf{Training dynamics of DesertFormer.}
    Loss converges steadily over 40 epochs with no sign of
    overfitting.
    mIoU improves rapidly in the first 10 epochs and plateaus
    near epoch~30, consistent with the cosine annealing schedule.}
  \label{fig:curves}
\end{figure*}

Figure~\ref{fig:curves} shows the loss and mIoU curves over the
40 training epochs.
Loss decreases monotonically from 1.27 (train) / 1.15 (validation)
to convergence near 0.99 / 1.05 respectively.
Validation mIoU improves rapidly in the first 10 epochs
(0.536~$\to$~0.607) and continues to plateau around epoch~30,
consistent with the cosine annealing schedule.

\subsection{Failure Analysis}
\label{sec:failure}

\subsubsection{Confusion Matrix}
\begin{figure*}[t]
  \centering
  \includegraphics[width=\linewidth]{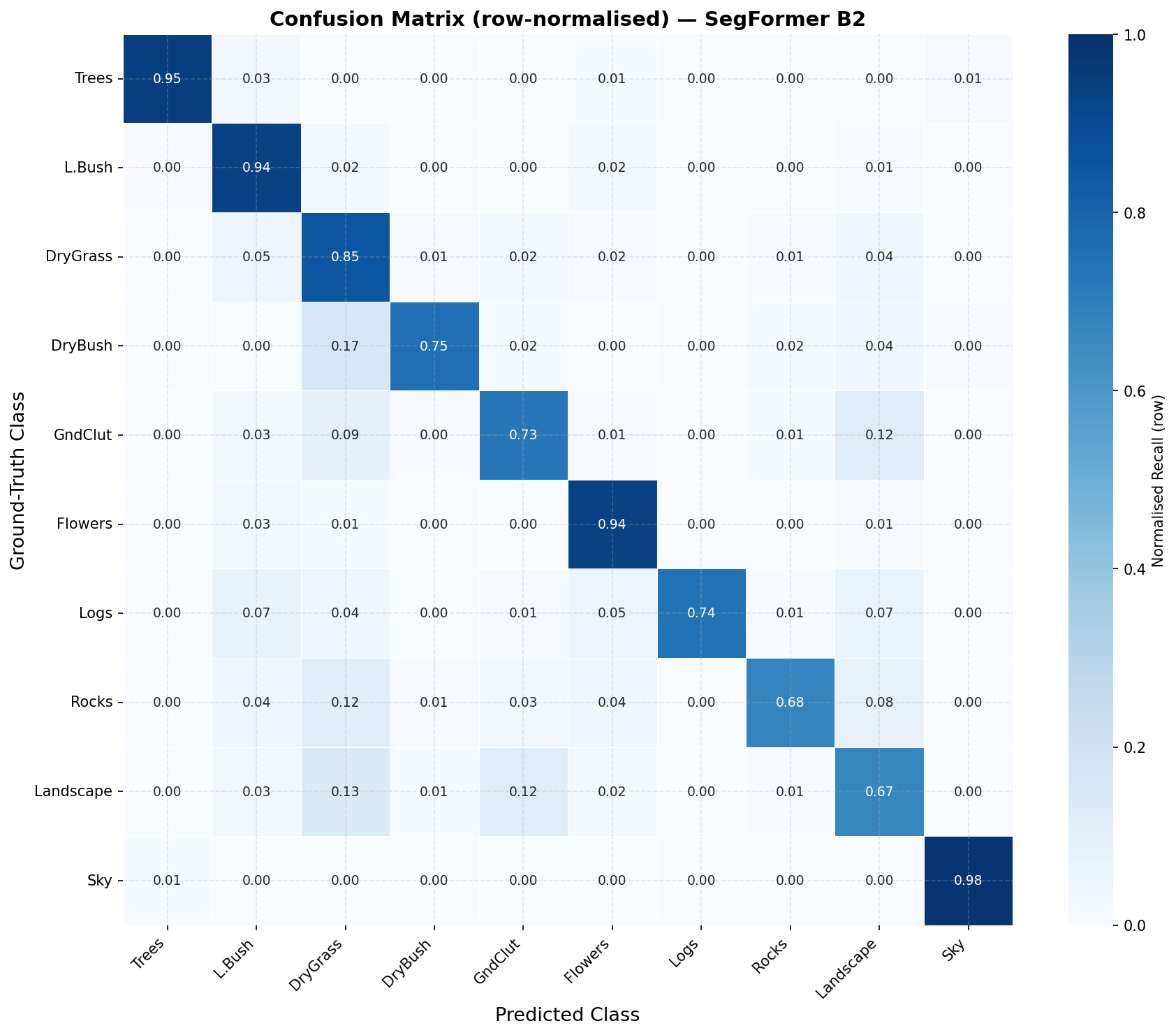}
  \caption{%
    \textbf{Row-normalised confusion matrix} on the validation set.
    Diagonal entries represent per-class recall.
    The off-diagonal hotspots at (Ground Clutter, Landscape) and
    (Dry Grass, Landscape) reveal the primary spectral confusion
    caused by similar sandy/earthy colours under desert lighting.}
  \label{fig:confusion_matrix}
\end{figure*}

The row-normalised confusion matrix
(Figure~\ref{fig:confusion_matrix}) reveals three dominant
misclassification pathways:
\begin{enumerate}[leftmargin=*,noitemsep]
  \item \textbf{Ground Clutter $\leftrightarrow$ Landscape}:
        15.67M confused pixels across the test set.
  \item \textbf{Dry Grass $\leftrightarrow$ Landscape}:
        12.17M confused pixels.
  \item \textbf{Dry Grass $\leftrightarrow$ Ground Clutter}:
        7.18M confused pixels.
\end{enumerate}
All three pairs share a common cause: in bright desert sunlight,
sandy ground, dried grass, and rocky debris exhibit nearly
identical hue and saturation values.
The spectral overlap creates an irreducible ambiguity that
purely appearance-based models struggle to resolve without
geometric or temporal context.

\subsubsection{Confidence Analysis}
Using a MC-Dropout uncertainty estimator~\cite{gal2016dropout}
over the 1{,}002 test images:
\begin{itemize}[leftmargin=*,noitemsep]
  \item \textbf{Global mean confidence}: 0.650
  \item \textbf{Uncertain pixels} (above entropy threshold): 34.0\%
  \item \textbf{High-uncertainty images}: 158 (15.8\%)
  \item \textbf{Well-predicted images}: 531 (53.0\%)
\end{itemize}
The hardest test image (\texttt{0000598.png}, difficulty score
0.430) contains 52.7\% uncertain pixels with simultaneous
Ground~Clutter / Landscape / Dry~Grass ambiguity in the foreground.

\subsection{Qualitative Analysis}

\begin{figure*}[t]
  \centering
  \begin{minipage}[t]{0.485\linewidth}
    \centering
    \includegraphics[width=\linewidth]{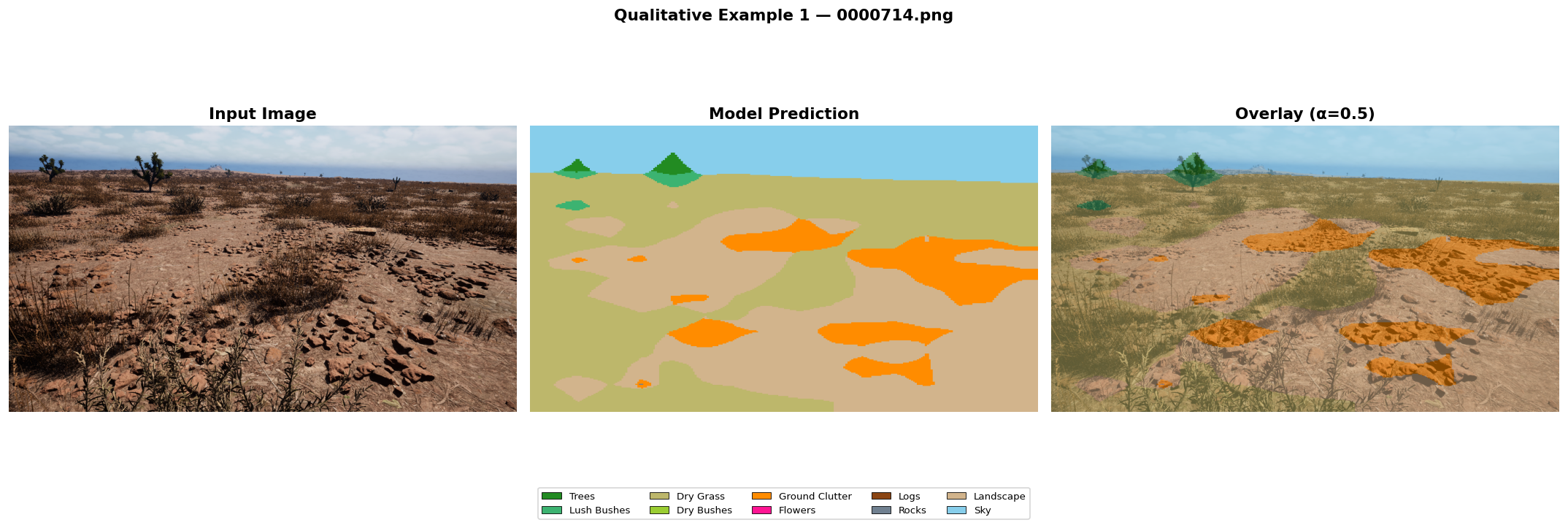}
    \par\vspace{2pt}
    {\footnotesize Ex.~1 --- open terrain, mixed vegetation}
  \end{minipage}
  \hfill
  \begin{minipage}[t]{0.485\linewidth}
    \centering
    \includegraphics[width=\linewidth]{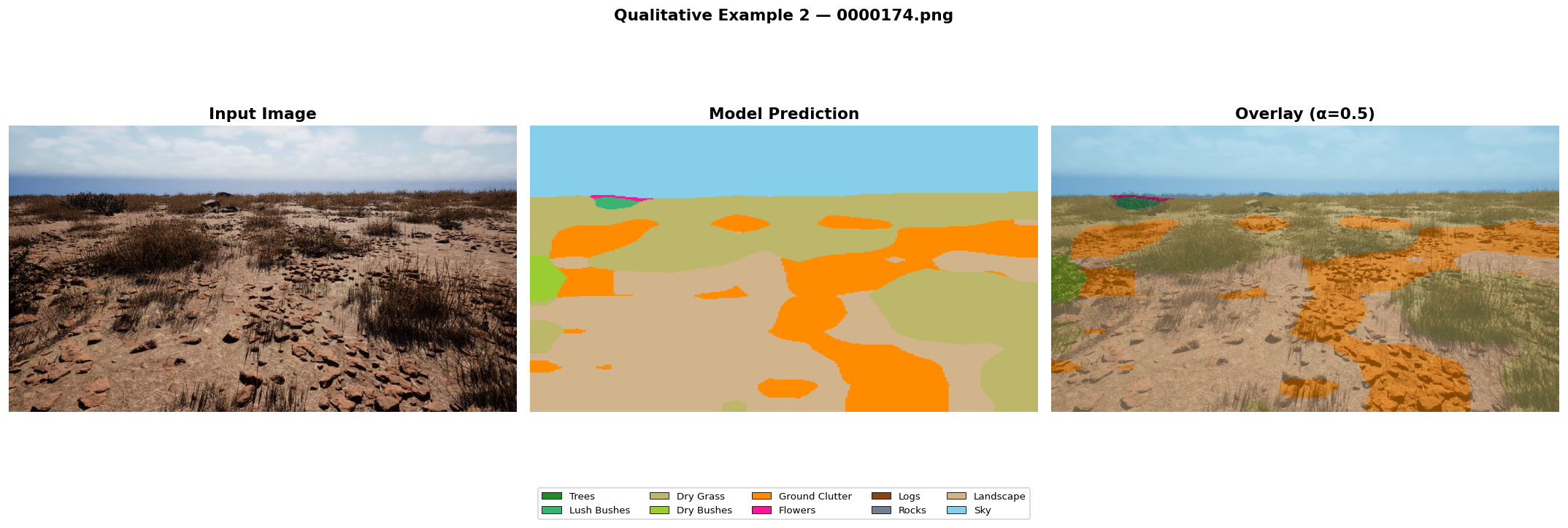}
    \par\vspace{2pt}
    {\footnotesize Ex.~2 --- dense sky + dry grass}
  \end{minipage}\\[6pt]
  \begin{minipage}[t]{0.485\linewidth}
    \centering
    \includegraphics[width=\linewidth]{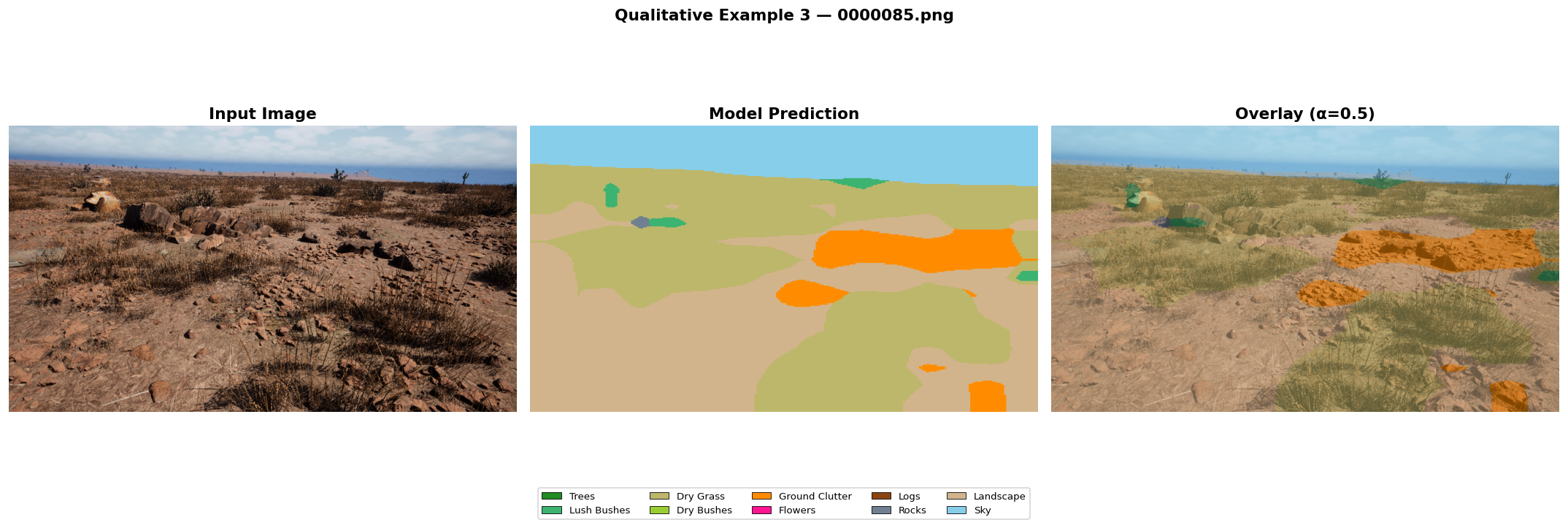}
    \par\vspace{2pt}
    {\footnotesize Ex.~3 --- rocky foreground}
  \end{minipage}
  \hfill
  \begin{minipage}[t]{0.485\linewidth}
    \centering
    \includegraphics[width=\linewidth]{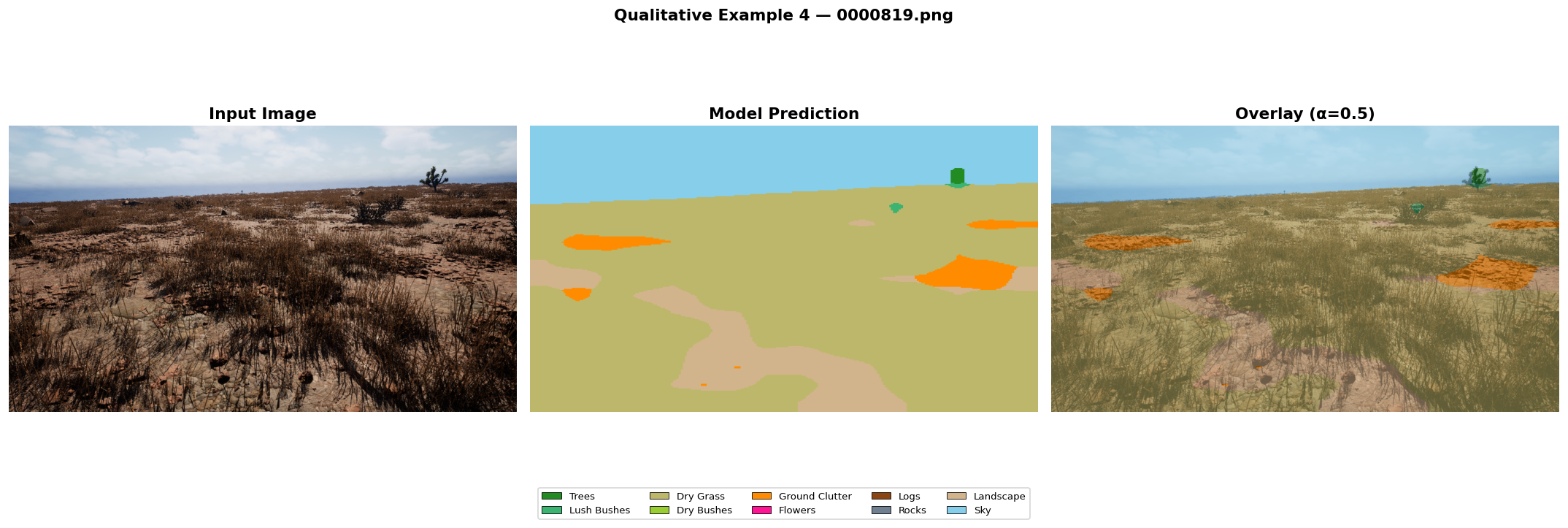}
    \par\vspace{2pt}
    {\footnotesize Ex.~4 --- landscape/ground clutter boundary}
  \end{minipage}\\[6pt]
  \begin{minipage}[t]{0.485\linewidth}
    \centering
    \includegraphics[width=\linewidth]{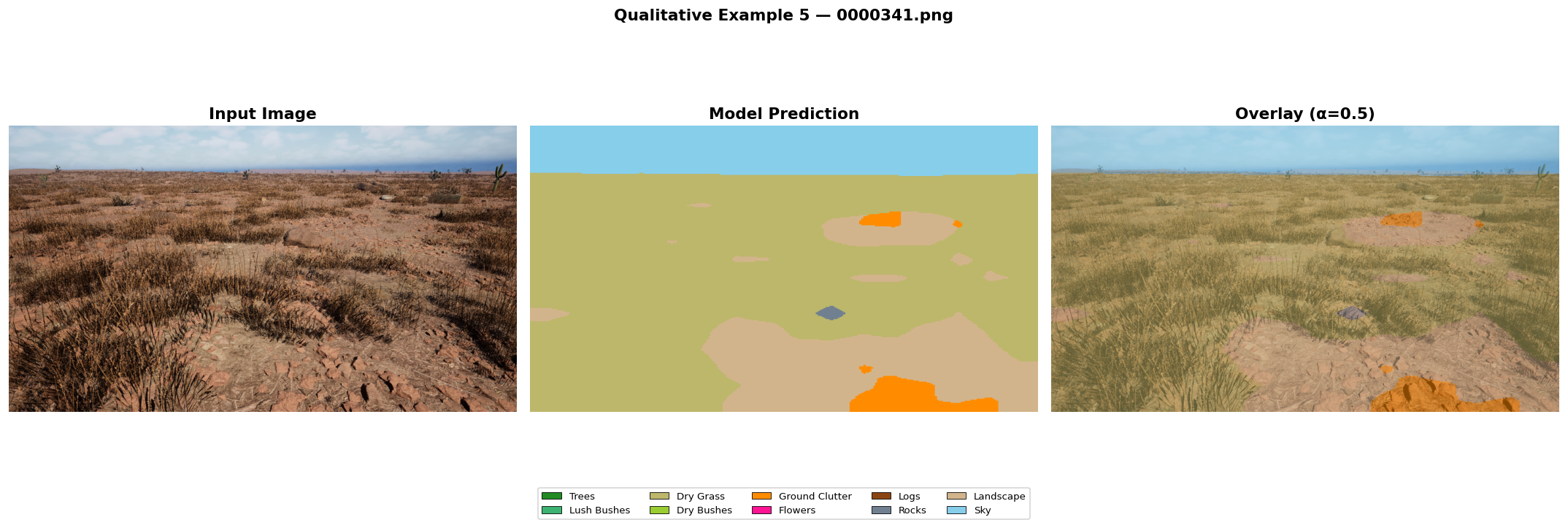}
    \par\vspace{2pt}
    {\footnotesize Ex.~5 --- lush bushes vs.\ dry bushes}
  \end{minipage}
  \hfill
  \begin{minipage}[t]{0.485\linewidth}
    \centering
    \includegraphics[width=\linewidth]{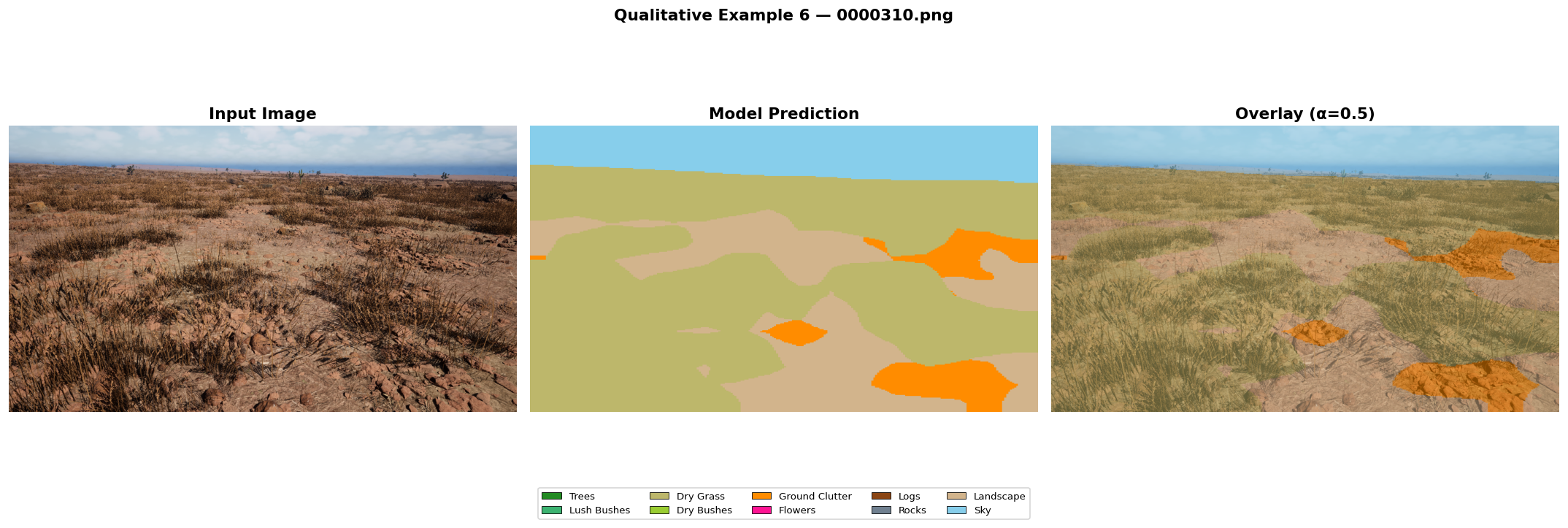}
    \par\vspace{2pt}
    {\footnotesize Ex.~6 --- wide open desert scene}
  \end{minipage}\\[6pt]
  \begin{minipage}[t]{0.485\linewidth}
    \centering
    \includegraphics[width=\linewidth]{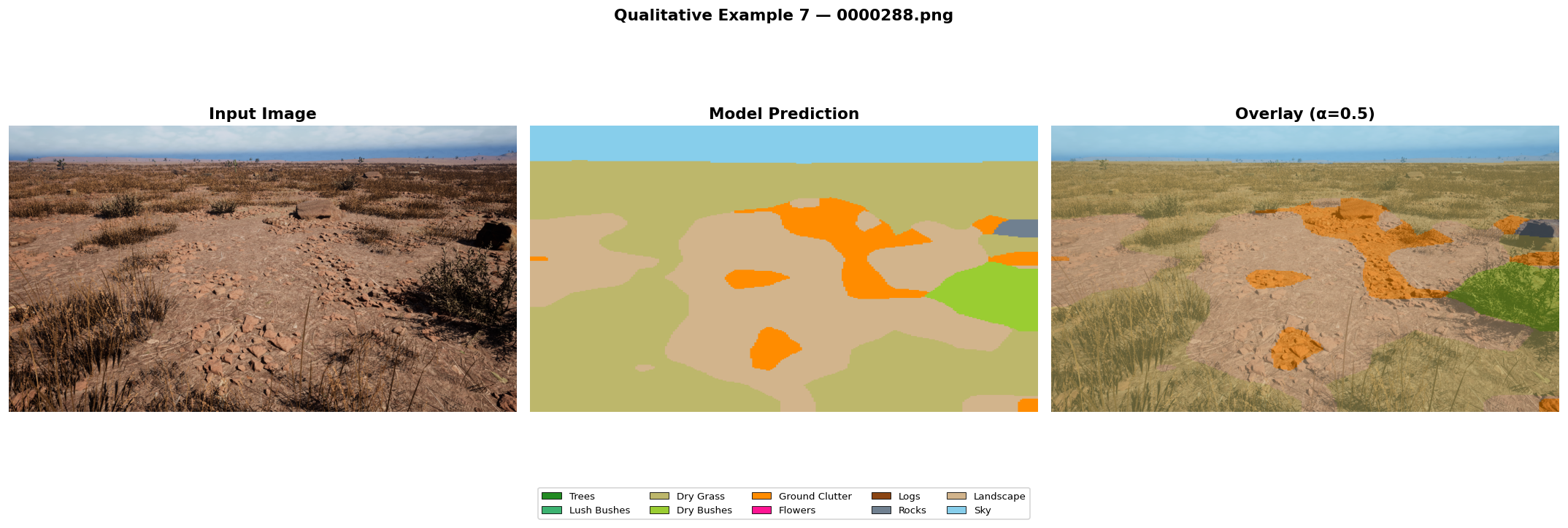}
    \par\vspace{2pt}
    {\footnotesize Ex.~7 --- trees and sky}
  \end{minipage}
  \hfill
  \begin{minipage}[t]{0.485\linewidth}
    \centering
    \includegraphics[width=\linewidth]{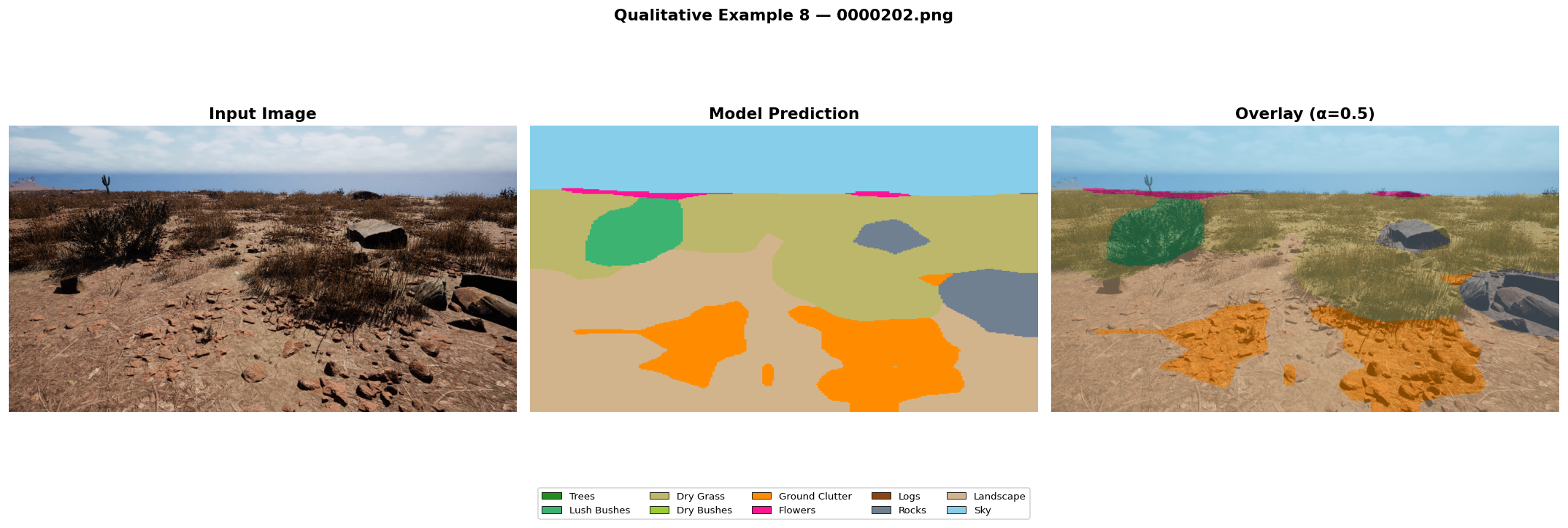}
    \par\vspace{2pt}
    {\footnotesize Ex.~8 --- challenging mixed terrain}
  \end{minipage}
  \caption{%
    \textbf{Qualitative results on eight randomly selected test images.}
    Each panel shows (left to right):
    \emph{input RGB} $|$ \emph{model prediction} $|$
    \emph{50\%-alpha overlay}.
    Class colours follow the palette in Figure~\ref{fig:bar_chart}.
    Vegetation (Trees, Dry Grass) and Sky are predicted with high
    spatial precision; the dominant failure mode is boundary
    ambiguity between Landscape, Dry Grass, and Ground Clutter
    in the mid-image transition zone.}
  \label{fig:qualitative}
\end{figure*}

Figure~\ref{fig:qualitative} presents eight representative
test-set examples (input RGB, prediction, 50\%-alpha overlay).
In success cases, Sky and Vegetation classes (Trees, Dry Grass)
are predicted with high spatial precision and well-aligned class
boundaries.
Failure cases predominantly occur in mid-image transition zones
between Landscape, Dry Grass, and Ground Clutter---a spatial
rather than categorical error, suggesting that Dense CRF
post-processing~\cite{krahenbuhl2011crf} could recover much of the
remaining accuracy.

\section{Discussion}
\label{sec:discussion}

\subsection{Applications}

\subsubsection{Terrain Safety Mapping}
Each semantic class can be mapped to a traversability cost:
\begin{itemize}[leftmargin=*,noitemsep]
  \item \textbf{Safe}: Landscape, Dry Grass, Sky
  \item \textbf{Caution}: Lush Bushes, Flowers, Ground Clutter
  \item \textbf{Obstacle}: Trees, Logs, Rocks, Dry Bushes
\end{itemize}
This three-level safety map feeds directly into costmap-based
path planners (e.g.\ ROS~2 Nav2) without additional processing.

\subsubsection{Autonomous Ground Robots}
The segmentation output drives a rover navigation simulator that
(i)~projects semantic masks into bird's-eye-view costmaps,
(ii)~highlights obstacle regions in the planned path, and
(iii)~suggests evasive waypoints in real-time.

\subsubsection{Edge Deployment}
The FastAPI inference server exposes a REST endpoint accepting
JPEG images and returning coloured segmentation masks with
per-pixel class probabilities.
INT8-quantised variants run at $\approx$4~FPS on a Jetson Nano,
suitable for survey drones or slow-moving ground vehicles.

\subsection{Limitations}

\paragraph{Class imbalance.}
Logs (0.07\%) and Dry Bushes (1.10\%) remain challenging despite
class-weighted loss and copy-paste augmentation.
Additional synthetic data generation or semi-supervised
pseudo-labelling of unlabelled images may help.

\paragraph{Domain specificity.}
The dataset covers a specific biome (arid/semi-arid desert).
Performance on temperate or tropical off-road environments has not
been evaluated and is expected to degrade without domain-adaptive
fine-tuning.

\paragraph{No temporal context.}
The model operates on individual frames and cannot exploit motion
continuity.
Sequential data from video-rate sensors would allow temporal models
to resolve boundary ambiguities that static appearance alone cannot.

\paragraph{Absence of depth information.}
Monocular RGB cannot distinguish visually similar surfaces at
different depths.
Fusion with LiDAR or stereo depth would directly address the
Ground~Clutter / Landscape confusion.

\subsection{Future Work}

\begin{itemize}[leftmargin=*,noitemsep]
  \item \textbf{Larger and more diverse datasets} spanning multiple
        desert biomes (Saharan, Arabian, Australian outback) to
        improve generalisation.
  \item \textbf{Multi-modal fusion} incorporating LiDAR point
        clouds or depth maps via cross-modal
        attention~\cite{zhang_cmx_2022,zhang_cmanet_2022},
        following CMX~\cite{zhang_cmx_2022} and RGB-D
        architectures~\cite{zhang_cmanet_2022,li_cmafnet_2023}.
  \item \textbf{Temporal segmentation} using Video
        Swin~\cite{liu2022videoswin} or recurrent feature banks
        to leverage inter-frame consistency.
  \item \textbf{Semi-supervised learning} exploiting unannotated
        off-road imagery to reduce annotation cost.
  \item \textbf{Real-time optimisation} via knowledge distillation
        into SegFormer~B0, targeting $\geq$30~FPS on embedded GPU.
  \item \textbf{3D terrain reconstruction} combining per-frame
        segmentation with Structure-from-Motion to produce
        semantically annotated point clouds for long-range planning.
\end{itemize}

\section{Conclusion}
\label{sec:conclusion}

We present \emph{DesertFormer}, a SegFormer~B2-based semantic
segmentation system for off-road desert terrain analysis.
On a purpose-built dataset of 4{,}176 images spanning ten terrain
classes, the pipeline achieves 64.4\% mIoU and 86.1\% pixel
accuracy---a 24.2 percentage point improvement over the DeepLabV3
baseline.
We demonstrate that combined CrossEntropy + Dice loss with
class-aware weighting and copy-paste augmentation for rare terrain
categories (Logs, Flowers, Dry Bushes) are critical for addressing
the severe pixel-frequency imbalance inherent to desert scenes.

Failure analysis reveals that the dominant confusion between
Ground~Clutter, Dry~Grass, and Landscape arises from spectral
similarity under desert lighting, an ambiguity that likely requires
geometric or temporal context to fully resolve.

The complete pipeline---training, evaluation, interactive
dashboard, CRF post-processing, and uncertainty estimation---is
released as open-source software to support the broader
autonomous-navigation research community.

\bibliographystyle{IEEEtran}
\bibliography{references}

\end{document}